\documentclass[10pt,twocolumn,letterpaper]{article}

\usepackage{cvpr}              
\usepackage{graphicx}
\usepackage{multirow}
\usepackage{makecell}
\usepackage{array}
\usepackage{adjustbox}
\usepackage{todonotes}

\definecolor{cvprblue}{rgb}{0.21,0.49,0.74}
\usepackage[pagebackref,breaklinks,colorlinks,allcolors=cvprblue]{hyperref}
\usepackage{nomencl} 
\makenomenclature 

\def\paperID{13379} 
\def\confName{CVPR}
\def\confYear{2025}

\title{Self-supervised Monocular Depth and Pose Estimation for Endoscopy with Latent Priors}

\author{Ziang Xu\textsuperscript{1}, Bin Li\textsuperscript{1}, Yang Hu\textsuperscript{1}, Chenyu Zhang\textsuperscript{1}, James East\textsuperscript{1}, Sharib Ali\textsuperscript{2}, Jens Rittscher\textsuperscript{1}\\
\textsuperscript{1}University of Oxford, \textsuperscript{2}University of Leeds\\
}

\begin{document}
\maketitle
\begin{abstract}

Accurate 3D mapping in endoscopy enables quantitative, holistic lesion characterization within the gastrointestinal (GI) tract, requiring reliable depth and pose estimation. However, endoscopy systems are monocular, and existing methods relying on synthetic datasets or complex models often lack generalizability in challenging endoscopic conditions. We propose a robust self-supervised monocular depth and pose estimation framework that incorporates a Generative Latent Bank and a Variational Autoencoder (VAE). The Generative Latent Bank leverages extensive depth scenes from natural images to condition the depth network, enhancing realism and robustness of depth predictions through latent feature priors. For pose estimation, we reformulate it within a VAE framework, treating pose transitions as latent variables to regularize scale, stabilize z-axis prominence, and improve x-y sensitivity. This dual refinement pipeline enables accurate depth and pose predictions, effectively addressing the GI tract’s complex textures and lighting. Extensive evaluations on SimCol and EndoSLAM datasets confirm our framework’s superior performance over published self-supervised methods in endoscopic depth and pose estimation.

\end{abstract}    
\section{Introduction}
\label{sec:intro}

Colorectal cancer (CRC) is a major global health concern, ranking as the third most common cancer worldwide and responsible for approximately 35\% of cancer-related deaths~\cite{biller2021diagnosis}. Accurate lesion mapping in colonoscopy is essential for effective CRC diagnosis and treatment but remains challenging due to its reliance on endoscopist expertise~\cite{bretthauer2022effect}.
In this context, 3D reconstruction offers a powerful, consistent method for precise lesion mapping and structural assessment, enhancing clinical accuracy~\cite{ma2019real}.

Effective 3D reconstruction in endoscopy relies on depth and pose estimation. Depth estimation provides spatial information, while pose estimation tracks camera orientation and movement through the GI tract. 
However, reliable ground truth for depth and pose are difficult to obtain due to the dynamic, constricted nature of the intestines. 
Additionally, endoscopy typically relies on monocular imaging, which limits depth perception and complicates 3D mapping in the GI tract’s variable and occluded environment.

\begin{figure}[!t]
    \centering
    \captionsetup{aboveskip=2pt, belowskip=-10pt} 
    \includegraphics[width=0.9\columnwidth]{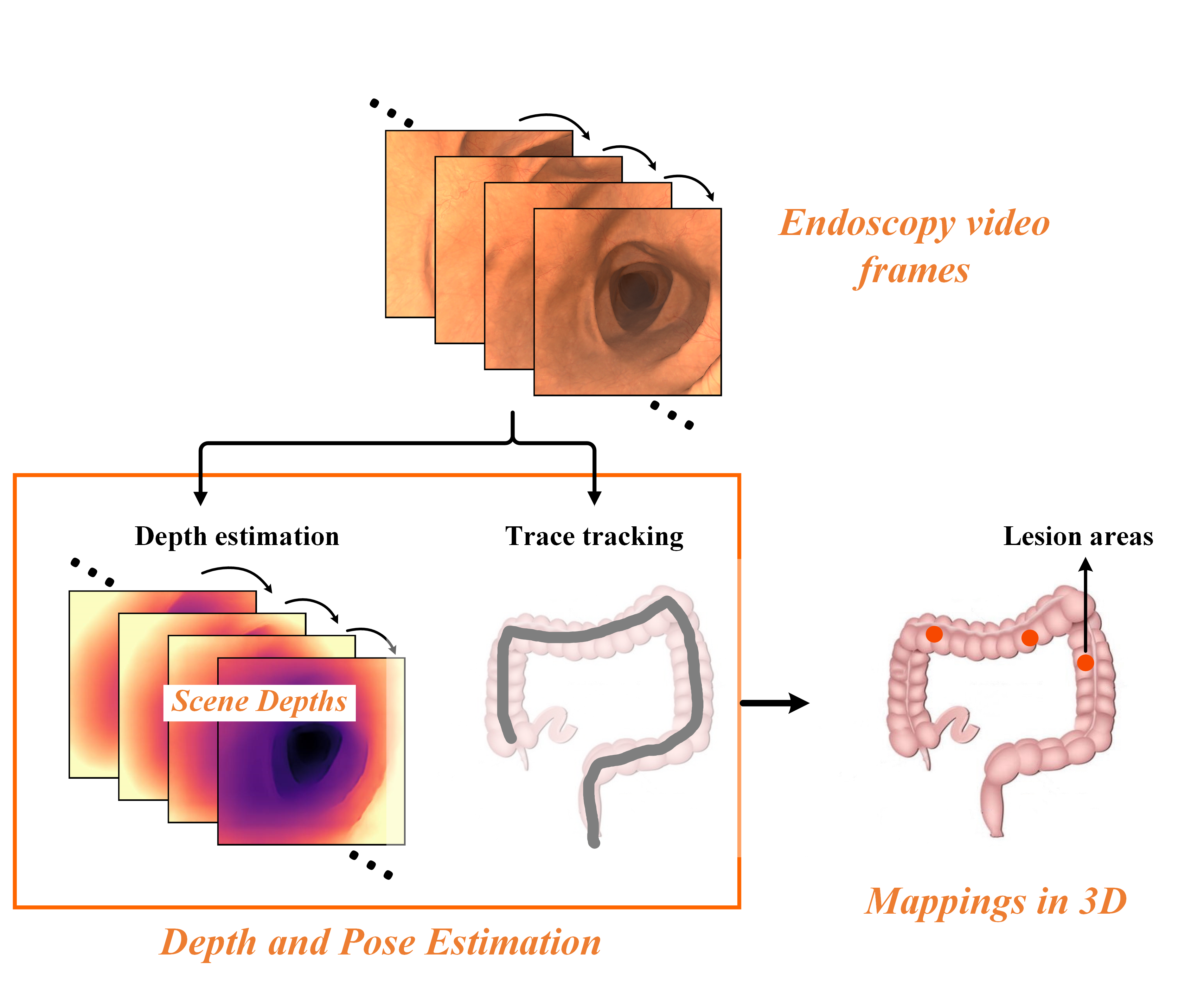}
    \caption{Workflow for 3D lesion mapping in endoscopy: depth estimation generates scene depth maps from monocular video frames, which, combined with pose trajectory, allow 3D reconstruction of the colon and precise lesion localization for improved diagnostics and surgical planning.}
    \label{fig:motivation}
\end{figure}
Recent research has addressed these challenges with synthetic datasets, self-supervised learning, and monocular approaches~\cite{hong20143d, yoon2022colonoscopic, bobrow2023colonoscopy, cheng2021depth, liu2023self}. 
Synthetic datasets enable robust training by simulating endoscopic conditions, while self-supervised learning leverages temporal and spatial cues to estimate depth and pose without manual labels.

While stereoscopic endoscopes are being developed they are not being used in routine practice. Hence, self-supervised monocular depth estimation is the preferred approach, framing depth estimation as a view synthesis problem and optimizing image reconstruction to avoid traditional depth labels. Monocular systems rely on a single view and require an additional pose network to infer motion, adding complexity to compensate for limited depth cues. 


Despite recent advances in monocular depth and pose estimation—such as Monodepth~\cite{godard2017unsupervised, godard2019digging}, Depth Anything~\cite{depthanything}, and DepthPro~\cite{Bochkovskii2024:arxiv}—current methods struggle with 3D colonoscopy due to the unique challenges of the endoscopic environment, including complex textures, erratic scene contrast, and reliance on synthetic data.
To address these issues, we propose a robust approach that incorporates \textit{generative latent priors} into a joint self-supervised framework for depth and pose estimation, conditioning predictions to overcome these challenges.

Our self-supervised backbone builds on Monodepth2’s principles~\cite{godard2019digging}, where a depth network estimates the current scene’s depth and a pose network predicts the relative pose transitions between frames. 
Then, reprojected adjacent frames are wrapped to predict the current frame, forming a self-supervised loop by comparing this prediction with the actual frame. 
However, Monodepth2’s depth and pose predictions are driven solely by reprojection consistency, lacking conditioning for realistic scene depth or camera poses.

To improve depth estimation, we incorporate a Generative Latent Bank trained with StyleGAN framework on depth images~\cite{karras2019style, chan2021glean}, conditioning the depth network with a prompting-based approach. Here, the depth encoder generates prompts processed by the pretrained latent bank, which incorporates prior knowledge of depth scenes, and the decoder assembles these atoms into the final depth estimate.

For pose estimation, we reframe the network as a Variational Autoencoder (VAE)~\cite{higgins2017beta, kingma2014semi} with the pose network as the encoder and the reprojection algorithm as the decoder, using pose estimates themselves as latent variables. 
This setup regularizes pose estimation, leveraging prior knowledge that colonoscopy pose transitions are generally gentle.

Our evaluations on SimCol~\cite{rau2019implicit} and EndoSlam~\cite{ozyoruk2021endoslam} datasets show improved performance over recent methods, and ablation studies validate the effectiveness of our proposed components.





\section{Related work}
\label{sec:related}


Most clinical endoscopy platforms are monocular, so our review focuses on key advances in monocular depth and pose estimation and their application in endoscopy. 
\textbf{Monocular depth estimation} is inherently challenging and ill-posed, as a single 2D image may correspond to multiple 3D scenes. Supervised deep learning approaches leverage accurate depth labels as supervision, enabling models to learn the relationship between RGB images and depth values. 
Eigen et al.~\cite{eigen2014depth} propose a dual-network model, where one network makes a global prediction, refined by a second network. 
A multi-scale approach~\cite{eigen2015predicting} further enhances depth estimation. 
Performance improvements have also come from novel loss functions, such as the Huber loss~\cite{laina2016deeper} and scale-invariant loss~\cite{lee2019big}, as well as by framing depth estimation as a multi-class classification~\cite{li2019deep} or regression task~\cite{fu2018deep}. 
Conditional Random Fields (CRFs) have been explored for post-processing~\cite{li2015depth,liu2015learning}.
Monocular video sequences can be used as supervised signals, but require the network to learn both depth and camera pose. 
To optimise model effects others have introduced additional constraints such uncertainty~\cite{poggi2020uncertainty,marsal2024monoprob,yang2020d3vo}, normal consistency~\cite{yang2018lego,yang1711unsupervised}, semantic segmentation~\cite{kumar2021syndistnet,jung2021fine,casser2019unsupervised}, and visual odometry~\cite{wang2018learning}. 
Remarkably, Monodepth2~\cite{godard2019digging} enhances model convergence speed and accuracy without additional constraints by optimizing minimum reprojection loss and introducing automatic masking to handle static objects. Several subsequent methods build on Monodepth2, including DualRefine~\cite{bangunharcana2023dualrefine}, MonoViT~\cite{zhao2022monovit}, and Lite-Mono~\cite{zhang2023lite}.

Real depth maps are challenging to obtain in endoscopy due to the need for specialized depth sensors and the variability in measurement quality~\cite{matthias20183d}. 
Synthetic data and self-supervised learning methods address this by reframing monocular depth estimation as a view synthesis problem, eliminating the need for ground truth depth data.
To improve depth and pose estimation in endoscopy, Mahmood et al.~\cite{mahmood2018deep} proposed a joint CNN-CRF framework trained on synthetic datasets with ground truth and adapted through adversarial training. 
Widya et al.~\cite{widya2021learning} introduced a generalized luminosity loss to balance depth and pose losses, enhancing model generalization across synthetic and real data. Generative approaches, such as those by Rau et al.~\cite{rau2019implicit} and Mahmood et al.~\cite{mahmood2019polyp}, employed GANs to generate depth maps, though direct GAN-generated maps often lack accuracy, and supervised learning is impractical given the need for extensive ground truth data in endoscopy.

Self-supervised methods, which eliminate the need for ground truth depth maps, have shown promise. Hwang et al.~\cite{hwang2021unsupervised} proposed a depth feedback network using self-supervised neighboring frame depth prediction with reconstruction error computation. 
Ozyoruk et al.~\cite{ozyoruk2021endoslam} combined a residual network with spatial attention and a luminance-aware loss to improve robustness under varying illumination and introduced the EndoSLAM dataset. 
Liu et al.~\cite{liu2023self} presented a self-supervised monocular depth estimation model with dual attention, using multi-scale structural similarity and $L_1$ losses to maintain luminance and color invariance.

\begin{figure*}[h!]
    \centering
    \captionsetup{belowskip=-12pt} 
    \includegraphics[width=0.8\textwidth]{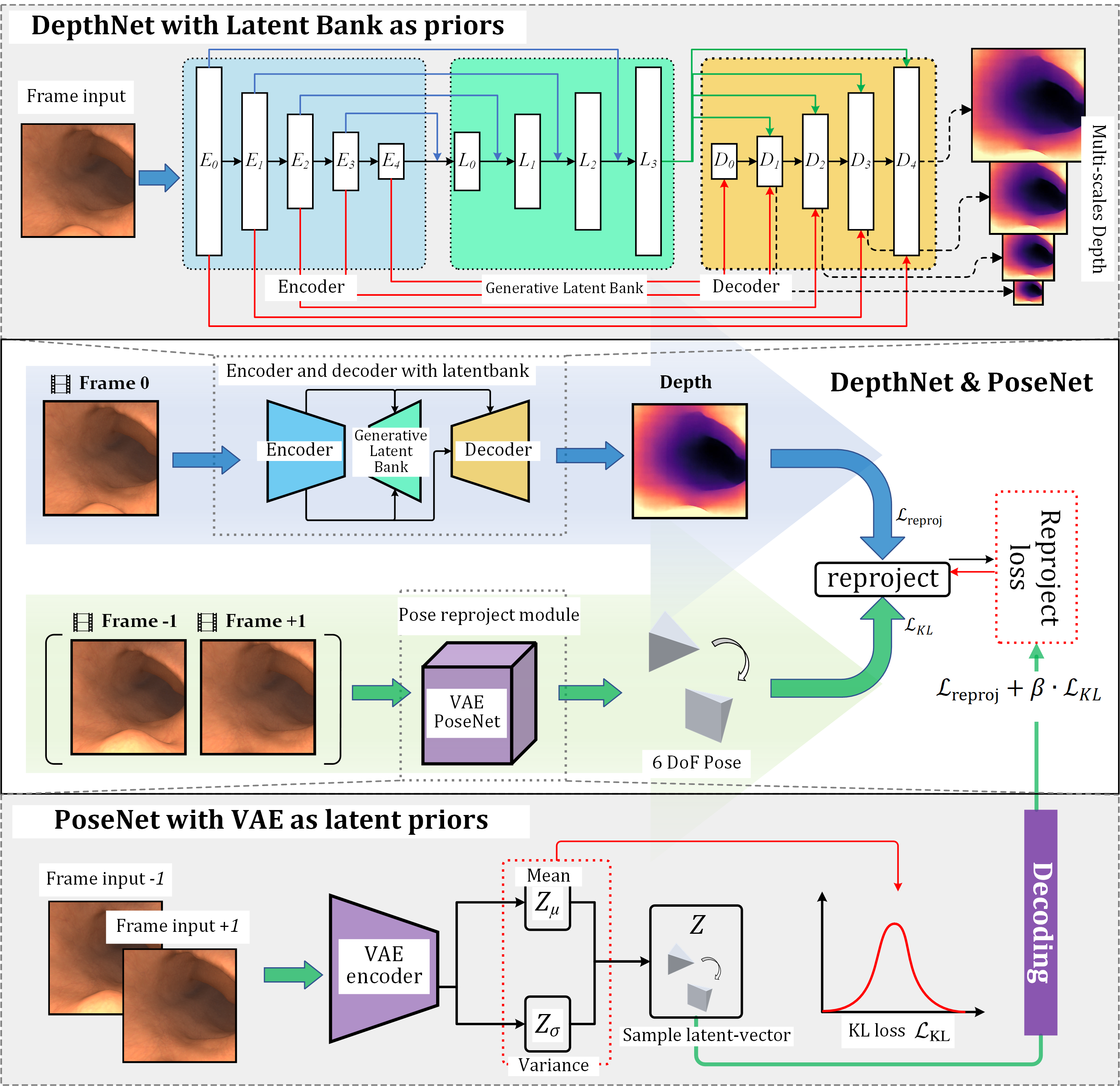} 
    \caption{Overview of the proposed method. The method consists of a depth estimation network with a pre-trained Generative Latent Bank and a VAE-constrained pose estimation network. The entire method is self-supervised training through subsequent reprojection as a supervision signal.}
    \label{fig:master}
\end{figure*}

\section{Methodology}
\label{sec:method}

Our model is designed to estimate depth and camera pose for reconstructing endoscopic movement traces and colon tract in a monocular setting. 
Briefly, our model comprises two main branches, DepthNet and PoseNet, which jointly predict depth and pose in a self-supervised framework, following the principles of Monodepth2~\cite{godard2019digging}.
Our architecture follows an encoding-decoding process (Fig~\ref{fig:master}). 
Specifically, DepthNet encodes the current frame's depth, while PoseNet encodes the camera poses of adjacent frames (frames +1 and -1). 
The encoded outputs are used in a reprojection step that warps adjacent RGB frames to reconstruct the current frame (frame 0). 
A reconstruction loss is then calculated between the warped output and the ground truth frame, completing the self-supervised learning loop.

To enhance the encoding process, we inject generative latent priors into both depth and pose branches. 
For depth encoding, we use a generative latent bank~\cite{chan2021glean} that provides structured depth “atoms,” guiding the network to produce realistic depth-map outputs. 
For pose encoding, we incorporate a variational autoencoder (VAE) where the pose network serves as the encoder, with predicted poses as latent variables and the reprojection process as the decoder. The VAE’s KL divergence term enforces smooth, scale-consistent pose transitions across XYZ axes.

The following sections detail each model component.

\subsection{DepthNet with Generative Latent Bank}
\subsubsection{Encoder and Decoder}
\label{sec:311}
Our DepthNet consists of an encoder $\mathbf{E_{dep}}$, a latent bank $\mathbf{L_{depth}}$, and a decoder $\mathbf{D_{dep}}$. The input RGB image, $\mathbf{x_i}$, is processed by the encoder, which comprises a series of downsampling and convolutional layers, producing a set of output feature maps at decreasing resolutions, $\mathbf{h_i} = \{h_i^{n-1}, \dots, h_i^0\}$, where $n$ represents the number of resolutions.
The output feature map at each resolution is passed to the latent bank. The structure of the latent bank follows the scheme of a StyleGAN generator~\cite{karras2019style}, which takes a latent vector and performs progressive upscaling using transpose convolutions. At each resolution, an additional latent vector is fused with the upscaled feature maps using Adaptive Instance Normalization (AdIN)~\cite{huang2017arbitrary}.

When functioning as a latent bank, the feature maps produced by the DepthNet encoder, i.e., $\mathbf{h_i} = \{h_i^{n-1}, \dots, h_i^0\}$, are treated as latent vectors and fused into the upscaling cascade at their corresponding resolutions, as shown in Fig~\ref{fig:master} top panel. The latent bank then generates output feature maps, $\mathbf{a_i} = \{a_i^{n-1}, \dots, a_i^0\}$, at each resolution, which are subsequently concatenated with the decoder $\mathbf{D_{dep}}$ at the corresponding resolutions.

The decoder $\mathbf{D_{dep}}$ receives both the feature maps returned by the latent bank, $\mathbf{a_i}$, and the feature maps from the encoder, $\mathbf{h_i}$, producing output depth predictions, $\mathbf{d_i} = \mathbf{D_{dep}}([h_i^k, a_i^k], k \in \{n-1, 0\}) = \{d_i^0, d_i^1, d_i^2, d_i^3\}$, at the four largest resolutions. These correspond to depth predictions at four different scales, as illustrated in Fig~\ref{fig:master} top panel.
The key idea is to leverage the information stored in the latent bank and reformulate depth prediction as a prompting process, where the encoder $\mathbf{E_{dep}}$ extracts features from the RGB images that serve as prompts to the pretrained latent bank $\mathbf{L_{depth}}$. The retrieved features from $\mathbf{L_{depth}}$ are then used in the decoding process to reconstruct depth maps.

By pretraining the latent bank $\mathbf{L_{depth}}$ with a large amount of scene depth data, we condition the depth generation process, encouraging the decoder to produce outputs that resemble genuine depth scenes. Next, we will elaborate on the pretraining process of the latent bank.


\subsubsection{Generative Latent Bank}
\textbf{Generative pretraining mode}.

\begin{figure}[t!]
    \centering
    \captionsetup{aboveskip=2pt, belowskip=-10pt} 
    \includegraphics[width=\columnwidth]{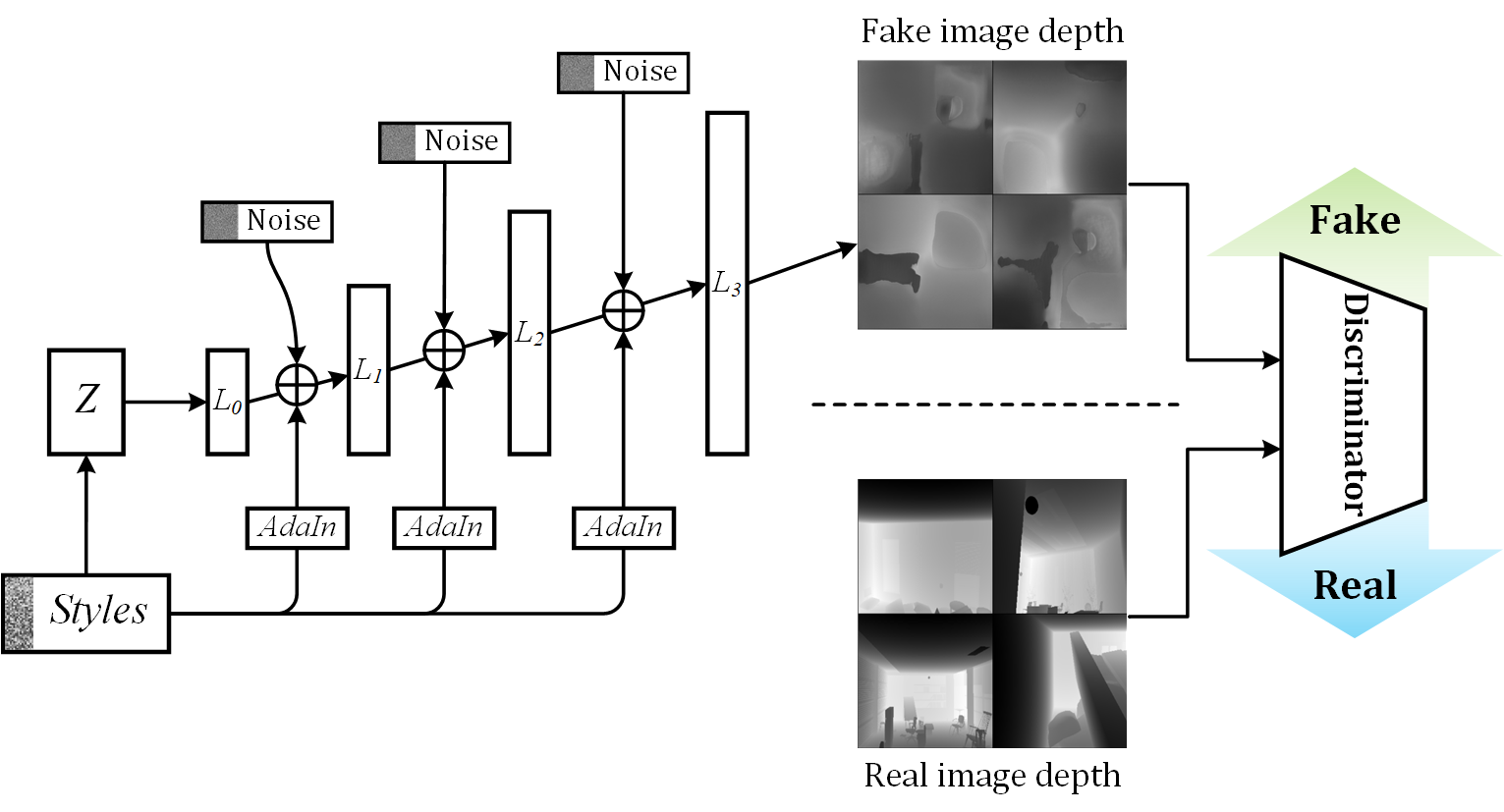}
    \caption{The pre-training process of Generative Latent Bank. A Gaussian latent vector is upsampled through transpose convolutions, with adaptive noise injections at each resolution, to produce variable depth maps. Trained in a GAN framework, the latent bank generates realistic depth maps, while a discriminator classifies real and synthetic maps to refine generation quality.}
    \label{fig:stylegan}
\end{figure}

The latent bank is a pretrained image generation network that follows a StyleGAN-like scheme~\ref{fig:stylegan}. The core idea is to amortize the depth prediction process by injecting prior knowledge of scene depths (e.g., from natural scenes or other sources where depth data is abundant and easier to obtain compared to endoscopy).

In each iteration of the generative pretraining, a random Gaussian latent vector is generated and passed through a series of upsampling blocks using transpose convolutions. At each resolution level, a new Gaussian noise vector is combined with the features via Adaptive Instance Normalization (AdIN), adding variability to the outputs. 
After upscaling to the target resolution, a final convolutional block reduces the channels to 1, producing the output depth map.

The discriminator $\mathbf{D_{gan}}$ is a standard residual CNN trained to classify real vs. synthetic depth maps generated by the generator (i.e., generative latent bank $\mathbf{L_{depth}}$). 
The discriminator takes in a depth map and outputs a binary classification score, where a score of 1 indicates a real depth map, and a score of 0 indicates a synthetic one. 

The latent bank $\mathbf{L_{depth}}$ and discriminator $\mathbf{D_{gan}}$ are trained in a GAN framework. The goal is to optimize the latent bank parameters $\theta_{LB}$ to generate realistic depth maps, while the discriminator parameters $\theta_{d}$ are optimized to accurately distinguish real from generated depth maps.

This is done by solving the following min-max problem:

\[
\min_{\theta_{LB}} \max_{\theta_{d}} \mathcal{L}_{GAN}(\theta_{LB}, \theta_{d})
\]

The GAN loss $\mathcal{L}_{GAN}$ consists of two components: the Wasserstein loss~\cite{arjovsky2017wasserstein}, which ensures a smooth measure of the distance between real and generated depth distributions, and a gradient penalty~\cite{gulrajani2017improved} that enforces the Lipschitz constraint on the critic. 

\begin{equation}
\begin{aligned}
\mathcal{L}_{GAN}(\theta_{LB}, \theta_{d}) = & \mathbb{E}_{x \sim p_{real}}[D_{gan}(x)] \\
& - \mathbb{E}_{z \sim p_{z}}[D_{gan}(L_{depth}(z))] \\
& + \gamma \cdot \mathbb{E}_{\hat{x} \sim p_{\hat{x}}}[(\|\nabla_{\hat{x}} D_{gan}(\hat{x})\|_2 - 1)^2]
\end{aligned}
\label{eq:gan_loss}
\end{equation}

\begin{itemize}
    \item $\mathbb{E}_{x \sim p_{real}}[D_{gan}(x)]$ is the expected output of the discriminator for real depth maps.
    \item $\mathbb{E}_{z \sim p_{z}}[D_{gan}(L_{depth}(z))]$ is the expected output of the discriminator for generated depth maps.
    \item $\gamma$ is the weight for the gradient penalty term.
    \item $\mathbb{E}_{\hat{x} \sim p_{\hat{x}}}[(\|\nabla_{\hat{x}} D_{gan}(\hat{x})\|_2 - 1)^2]$ is the gradient penalty term to enforce the Lipschitz constraint~\cite{gulrajani2017improved}.
\end{itemize}

The latent bank parameters $\theta_{LB}$ are optimized to minimize the adversarial loss with gradient penalty, while the discriminator parameters $\theta_{d}$ are optimized to maximize its ability to classify real vs. generated samples. The training process alternates between minimizing the generator’s loss and maximizing the discriminator’s classification accuracy.

\textbf{Latent bank mode}.
After the latent bank $\mathbf{L_{depth}}$ is trained to produce realistic scene depth outputs, its weights are frozen, and it is integrated into the encoder-decoder architecture discussed in section~\ref{sec:311}. Note that a new set of trainable Adaptive Instance Normalization (AdIN) blocks is initialized and optimized during the self-supervised training of the depth encoder $\mathbf{E_{dep}}$ and the decoder $\mathbf{D_{dep}}$.
The output of the decoder $\mathbf{d_i} = \mathbf{D_{dep}}(\mathbf{a_i}, \mathbf{h_i})$ will be fed to the reprojection algorithm to reconstruct the input RGB frame and the training process of $\mathbf{D_{dep}}$ and $\mathbf{E_{dep}}$ will be discussed in section~\ref{sec:33}.

\subsection{VAE-constrained PoseNet}
For generating pose predictions, we utilize a pose encoder $\mathbf{E_{pos}}$ that takes an RGB input and generates a pose estimation, represented as a vector of shape $\mathbb{R}^{6 \times 1}$, consisting of six pose parameters. The pose estimation is computed separately for both the previous frame (\(-1\)) and the subsequent frame (\(+1\)). After obtaining the pose estimations from $\mathbf{E_{pos}}$, these estimated poses are input to the reprojection algorithm along with the depth output $d_i$ from DepthNet.
The reprojection algorithm, serving as the decoder in this context, takes the estimated poses and depth maps and warps the RGB images of the \(-1\) and \(+1\) frames to produce reprojected RGB images at the current frame. The reprojected images are then compared with the ground truth RGB image of the current frame using MSE loss, ensuring spatial temporal consistency and completing the self-supervision.

The key difference in this approach is that we treat the output of PoseNet, which represents the relative pose differences between adjacent frames (-1 and +1) to the current frame, as latent variables in a Variational Autoencoder (VAE). Specifically, the reprojection algorithm acts as the decoder in this setup, constraining these relative pose parameters by enforcing a KL divergence between the predicted pose differences and a Gaussian prior. 
This regularization suppresses the prominence in z-axis movements while improves the relative sensitivity of x-y pose chamges between adjacent frames, reflecting prior knowledge that endoscopic movements are typically smooth to minimize potential damage to the GI tract.

Empirically, we demonstrate that using VAE regularization on the scales of poses along the \(x\), \(y\), and \(z\) axes results in a more coherent and sensible reconstruction of the endoscopic movement trace. The overall optimization objective consists of two main components:

\textbf{Reprojection Loss}: This is computed as the mean squared error (MSE) between the ground truth RGB image $\mathbf{x}_{\text{gt}}$ at the current frame and the reprojected RGB image $\mathbf{x}_{\text{reproj}}$, obtained by warping the adjacent frames using the estimated poses and depth (explained in section~\ref{sec:33}):
\[
\mathcal{L}_{\text{reproj}} = \| \mathbf{x}_{\text{gt}} - \mathbf{x}_{\text{reproj}} \|_2^2.
\]

\textbf{KL Divergence Regularization}: The KL divergence between the predicted pose parameters and a Gaussian prior, which smooths the changes in pose estimates:
\[
\mathcal{L}_{\text{KL}} = \text{KL}\left( q(\mathbf{z_{pos}}) \| \mathcal{N}(0, \mathbf{I}) \right),
\]
where $q(\mathbf{z_{pos}})$ represents the distribution of pose parameters estimated by PoseNet, and $\mathcal{N}(0, \mathbf{I})$ denotes a standard Gaussian prior.

\textbf{Total Loss} for the VAE-constrained PoseNet:
\[
\mathcal{L}_{\text{total}} = \mathcal{L}_{\text{reproj}} + \beta \cdot \mathcal{L}_{\text{KL}},
\]
where $\beta$ controls the weight for KL divergence.

\begin{table*}[!t]
\centering
\captionsetup{aboveskip=2pt, belowskip=-10pt} 
\tiny
\renewcommand{\arraystretch}{1.2}
\begin{adjustbox}{width=\textwidth} 
\begin{tabular}{l|l|c|ccccc|ccc}
\hline
\multirow{2}{*}{\textbf{Evaluation}} & \multirow{2}{*}{\textbf{Method}} & \multirow{2}{*}{\textbf{Year}} & \multicolumn{5}{c|}{\textbf{Depth Error (}\(\downarrow\)\textbf{) (in cm)}} & \multicolumn{3}{c}{\textbf{Depth Accuracy (}\(\uparrow\)\textbf{)}} \\ 
 & & & \textbf{Abs Rel} & \textbf{Sq Rel} & \textbf{RMSE} & \textbf{RMSE log} & \textbf{$L_1$ error} & \(\delta < 1.25\) & \(\delta < 1.25^2\) & \(\delta < 1.25^3\) \\
\hline
\hline
\multirow{6}{*}{\textbf{Simcol-I, II}}
&Monodepth2 \cite{godard2019digging} & 2019 & 0.151 & 0.209 & 0.709 & 0.200 & 0.443 & 0.837 & 0.946 & 0.979 \\
&MonoViT \cite{zhao2022monovit} & 2022 & 0.138 & 0.234 & 0.715 & 0.157 & 0.423 & 0.863 & 0.952 & 0.952 \\
&DualRefine \cite{bangunharcana2023dualrefine} & 2023 & 0.133 & 0.289 & 0.717 & 0.152 & 0.425 & 0.871 & 0.944 & 0.974 \\
&Lite-Mono \cite{zhang2023lite} & 2023 & 0.126 & 0.184 & 0.640 & 0.144 & 0.358 & 0.886 & 0.969 & 0.990 \\
&Depth Pro \cite{Bochkovskii2024:arxiv} & 2024 & 0.174 & 0.137 & 0.708 & 0.211 & 0.479 & 0.734 & 0.949 & 0.988 \\
&\textbf{Ours} & 2024 & \textbf{0.125} & \textbf{0.110} & \textbf{0.614} & \textbf{0.126} & \textbf{0.347} & \textbf{0.937} & \textbf{0.983}  & \textbf{0.996}\\
\hline
\multirow{6}{*}{\textbf{Simcol-III}} 
&Monodepth2 \cite{godard2019digging} & 2019 & 0.135 & 0.137 & 0.624 & 0.187 & 0.412 & 0.840 & 0.937 & 0.957 \\
&MonoViT \cite{zhao2022monovit} & 2022 & 0.128 & 0.125 & 0.595 & 0.169 & 0.400 & 0.858 & 0.957 & 0.981 \\
&DualRefine \cite{bangunharcana2023dualrefine} & 2023 & 0.115 & 0.177 & 0.577 & 0.159 & 0.392 & 0.889 & 0.947 & 0.966 \\
&Lite-Mono \cite{zhang2023lite} & 2023 & 0.121 & 0.091 & 0.539 & 0.158 & 0.368 & 0.872 & 0.966 & 0.987 \\
&Depth Pro \cite{Bochkovskii2024:arxiv} & 2024 & 0.178 & 0.150 & 0.757 & 0.216 & 0.547 & 0.704 & 0.948 & 0.990 \\
&\textbf{Ours} & 2024 & \textbf{0.110} & \textbf{0.081} & \textbf{0.505} & \textbf{0.130} & \textbf{0.327} & \textbf{0.924} & \textbf{0.983}  & \textbf{0.991}\\
\hline
\multirow{5}{*}{\shortstack{\textbf{EndoSLAM} \\ \textbf{Colon}}}
 &Monodepth2 \cite{godard2019digging} & 2019 & 0.457 & 1.166 & 0.909 & 0.466 & 0.377 & 0.516 & 0.865 & 0.913 \\
 &MonoViT \cite{zhao2022monovit} & 2022 & 0.385 & 0.602 & 0.682 & 0.413 & 0.328 & \textbf{0.579} & 0.858 & 0.929 \\
 &DualRefine \cite{bangunharcana2023dualrefine} & 2023 & 0.423 & 0.469 & 0.643 & 0.439 & 0.404 & 0.479 & 0.749 & 0.865 \\
 &Lite-Mono \cite{zhang2023lite} & 2023 & 0.404 & 0.411 & 0.596 & 0.426 & 0.389 & 0.421 & 0.421 & 0.887 \\
 &\textbf{Ours} & 2024 & \textbf{0.351} & \textbf{0.314} & \textbf{0.540} & \textbf{0.387} & \textbf{0.239} & 0.566 & \textbf{0.866}  & \textbf{0.977}\\
\hline
\multirow{5}{*}{\shortstack{\textbf{EndoSLAM} \\ \textbf{Small Intestine}}}
 &Monodepth2 \cite{godard2019digging} & 2019 & 0.552 & 0.903 & 0.888 & 0.536 & 0.527 & 0.398 & 0.659 & 0.813 \\
 &MonoViT \cite{zhao2022monovit} & 2022 & 0.422 & 0.442 & 0.632 & 0.441 & 0.406 & 0.417 & 0.722 & 0.879 \\
 &DualRefine \cite{bangunharcana2023dualrefine} & 2023 & 0.423 & 0.495 & 0.617 & 0.442 & 0.403 & 0.418 & 0.704 & 0.875 \\
 &Lite-Mono \cite{zhang2023lite} & 2023 & 0.369 & 0.356 & 0.533 & 0.393 & 0.324 & 0.501 & 0.810 & 0.913 \\
 &\textbf{Ours} & 2024 & \textbf{0.336} & \textbf{0.296} & \textbf{0.503} & \textbf{0.376} & \textbf{0.204} & \textbf{0.648} & \textbf{0.648}  & \textbf{0.964}\\
 \hline
\end{tabular}
\end{adjustbox}
\caption{Comparison of depth estimation methods across various metrics on the SimCol and EndoSLAM dataset.}
\label{tab:depth}
\end{table*}

\subsection{Self-supervised Reprojection}
\label{sec:33}
The reprojection algorithm takes the depth map $d_i$ (output from DepthNet for the current frame), the RGB images of the adjacent frames ($\mathbf{x}_{-1}$ and $\mathbf{x}_{+1}$), and the estimated pose differences $\mathbf{z_{pos,-1}}$ and $\mathbf{z_{pos,+1}}$ (outputs of PoseNet) that specify the relative pose differences between the current frame and the adjacent frames. The algorithm uses these inputs to warp and interpolate the RGB images of \(-1\) and \(+1\) frames, generating the reprojected RGB image of the current frame, $\mathbf{I}_{\text{reproj}}$. The specific details of the reprojection algorithm follow the implementation in Monodepth2~\cite{godard2019digging}.

\textbf{Optimization Objective}  
We define the total optimization objective as minimizing the discrepancy between the current frame's RGB image and the reprojected image, while regularizing the pose estimates with a KL divergence term. The overall loss function for training all sub-networks—DepthNet (with $\mathbf{E_{dep}}$ and $\mathbf{D_{dep}}$) and PoseNet (with $\mathbf{E_{pos}}$)—is given by:

\begin{equation}
\begin{aligned}
    \mathcal{L}_{\text{total}}(\theta_{\text{dep}}, \theta_{\text{pos}}) = & \ \left\|\mathbf{x}_{0} - \text{Reproj}(\mathbf{x}_{-1}, \mathbf{z_{pos,-1}}, \mathbf{d_i})\right\|_2^2 \\
    & + \left\|\mathbf{x}_{0} - \text{Reproj}(\mathbf{x}_{+1}, \mathbf{z_{pos,+1}}, \mathbf{d_i})\right\|_2^2 \\
    & + \beta \cdot \text{KL}\left( q(\mathbf{z_{pos}}) \| \mathcal{N}(0, \mathbf{I}) \right)
\end{aligned}
\label{eq:total_loss}
\end{equation}

\begin{itemize}
    \item $\mathbf{x}_{0}$ denotes the current RGB frame.
    \item $\text{Reproj}(\cdot)$ represents the reprojection operation that uses the pose differences and depth estimates.
    \item $\mathbf{z_{pos,-1}}$ and $\mathbf{z_{pos,+1}}$ represent the estimated pose differences between the \(-1\) and \(+1\) frames, respectively, and the current frame.
    \item $\beta$ is a weighting factor for the KL divergence term.
    \item $q(\mathbf{z_{pos}})$ denotes the distribution of the estimated poses.
\end{itemize}

\section{Experiments}
\label{sec:experiment}

\subsection{Datasets}

\begin{table}[h]
\centering
\captionsetup{aboveskip=2pt, belowskip=-12pt} 
\tiny
\renewcommand{\arraystretch}{1.2}
\begin{adjustbox}{width=\columnwidth} 
\begin{tabular}{l|l|c|cc}
\hline
\multirow{2}{*}{\textbf{Evaluation}} & \multirow{2}{*}{\textbf{Method}} & \multirow{2}{*}{\textbf{Year}} & \multicolumn{2}{c}{\textbf{Pose Error (}\(\downarrow\)\textbf{)}} \\ 
 & & & \textbf{RTE} & \textbf{ROT} \\ 
\hline
\hline
\multirow{5}{*}{\textbf{Simcol-I, II}}
&Monodepth2 \cite{godard2019digging} & 2019 & 0.130 & 0.752 \\
&MonoViT \cite{zhao2022monovit} & 2022 & 0.108 & 0.708 \\
&DualRefine \cite{bangunharcana2023dualrefine} & 2023 & 0.109 & 0.697 \\
&Lite-Mono \cite{zhang2023lite} & 2023 & 0.110 & 0.715 \\
&\textbf{Ours} & 2024 & \textbf{0.089} & \textbf{0.671} \\
\hline
\multirow{5}{*}{\textbf{Simcol-III}}
&Monodepth2 \cite{godard2019digging} & 2019 & 0.361 & 0.932 \\
&MonoViT \cite{zhao2022monovit} & 2022 & 0.344 & 1.179 \\
&DualRefine \cite{bangunharcana2023dualrefine} & 2023 & 0.326 & 0.911 \\
&Lite-Mono \cite{zhang2023lite} & 2023 & 0.319 & 0.933 \\
&\textbf{Ours} & 2024 & \textbf{0.296} & \textbf{0.789} \\
\hline
\multirow{5}{*}{\shortstack{\textbf{EndoSLAM Colon,} \\ \textbf{Small Intestine}}}
&Monodepth2 \cite{godard2019digging} & 2019 & 0.561 & 0.887 \\
&MonoViT \cite{zhao2022monovit} & 2022 & 0.557 & 0.744 \\
&DualRefine \cite{bangunharcana2023dualrefine} & 2023 & 0.469 & 0.791 \\
&Lite-Mono \cite{zhang2023lite} & 2023 & 0.416 & 0.766 \\
&\textbf{Ours} & 2024 & \textbf{0.334} & \textbf{0.719} \\
\hline
\multirow{5}{*}{\shortstack{\textbf{EndoSLAM Colon-IV,} \\ \textbf{Traj-I}}}
&Monodepth2 \cite{godard2019digging} & 2019 & 0.025 & 0.228 \\
&MonoViT \cite{zhao2022monovit} & 2022 & 0.022 & 0.211 \\
&DualRefine \cite{bangunharcana2023dualrefine} & 2023 & 0.020 & 0.206 \\
&Lite-Mono \cite{zhang2023lite} & 2023 & 0.016 & 0.169 \\
&\textbf{Ours} & 2024 & \textbf{0.006} & \textbf{0.129} \\
\hline

\end{tabular}
\end{adjustbox}
\caption{Comparison of pose estimation methods across various metrics. Note that EndoSLAM Colon-IV,Traj-I contains real endoscopy video and ground truth camera poses measured from porcine GI organs.}
\label{tab:pose}
\end{table}

\begin{figure*}[!t]
    \centering
    \captionsetup{aboveskip=2pt, belowskip=-2pt} 
    \includegraphics[width=0.75\textwidth]{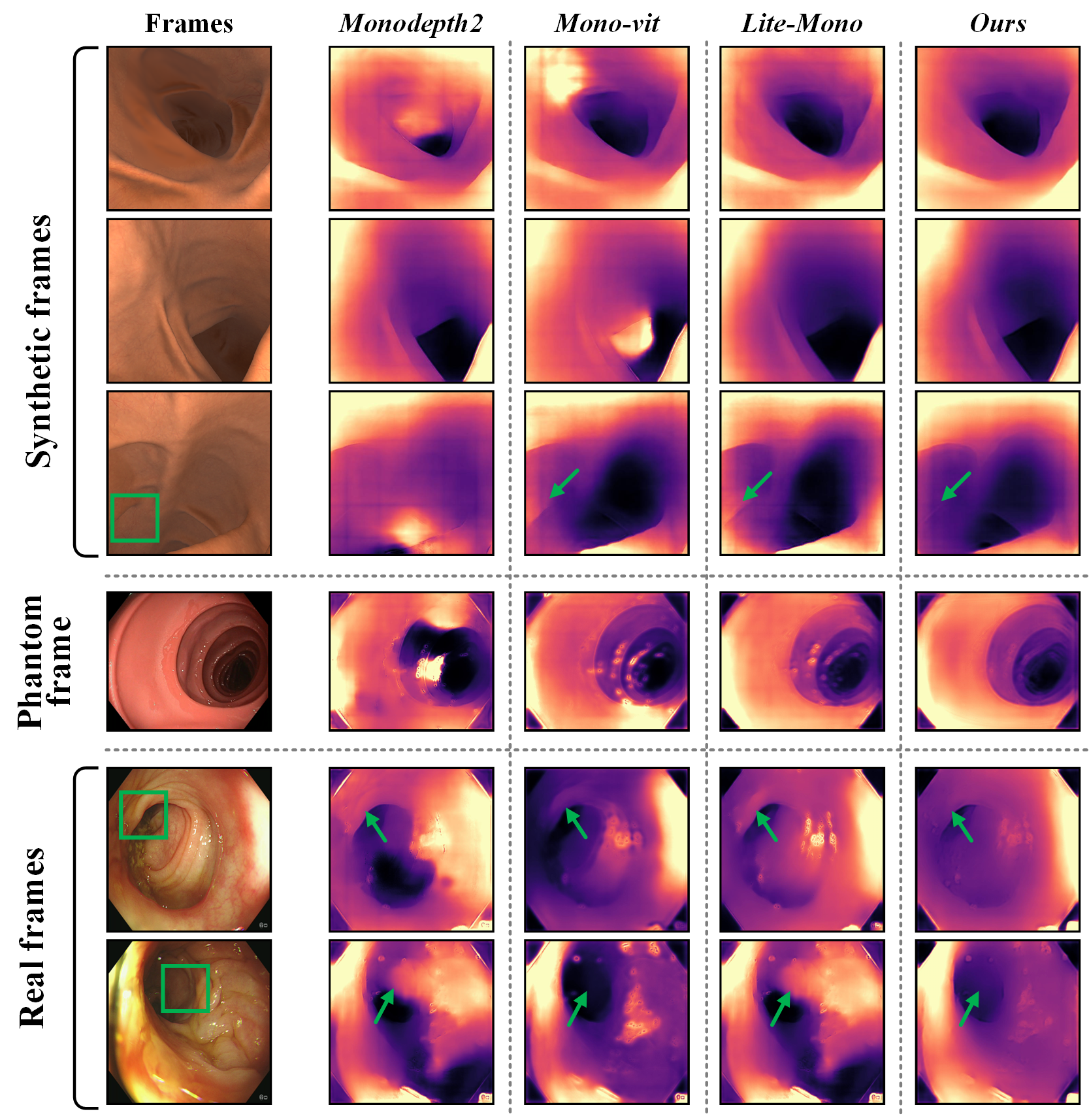} 
    \caption{Qualitative results of depth estimation. Here are some depth maps generated by Monodepth2~\cite{godard2019digging}, MonoVit~\cite{zhao2022monovit}, Lite-Mono~\cite{zhang2023lite} and Ours. Our method demonstrates superior performance, particularly on challenging phantom and real frames, where complex textures and lighting variations pose significant challenges for depth and pose estimation.}
    \label{fig:result_depth}
\end{figure*}

\begin{figure}[h]
    \centering
    \captionsetup{aboveskip=2pt, belowskip=-10pt} 
    \includegraphics[width=\columnwidth]{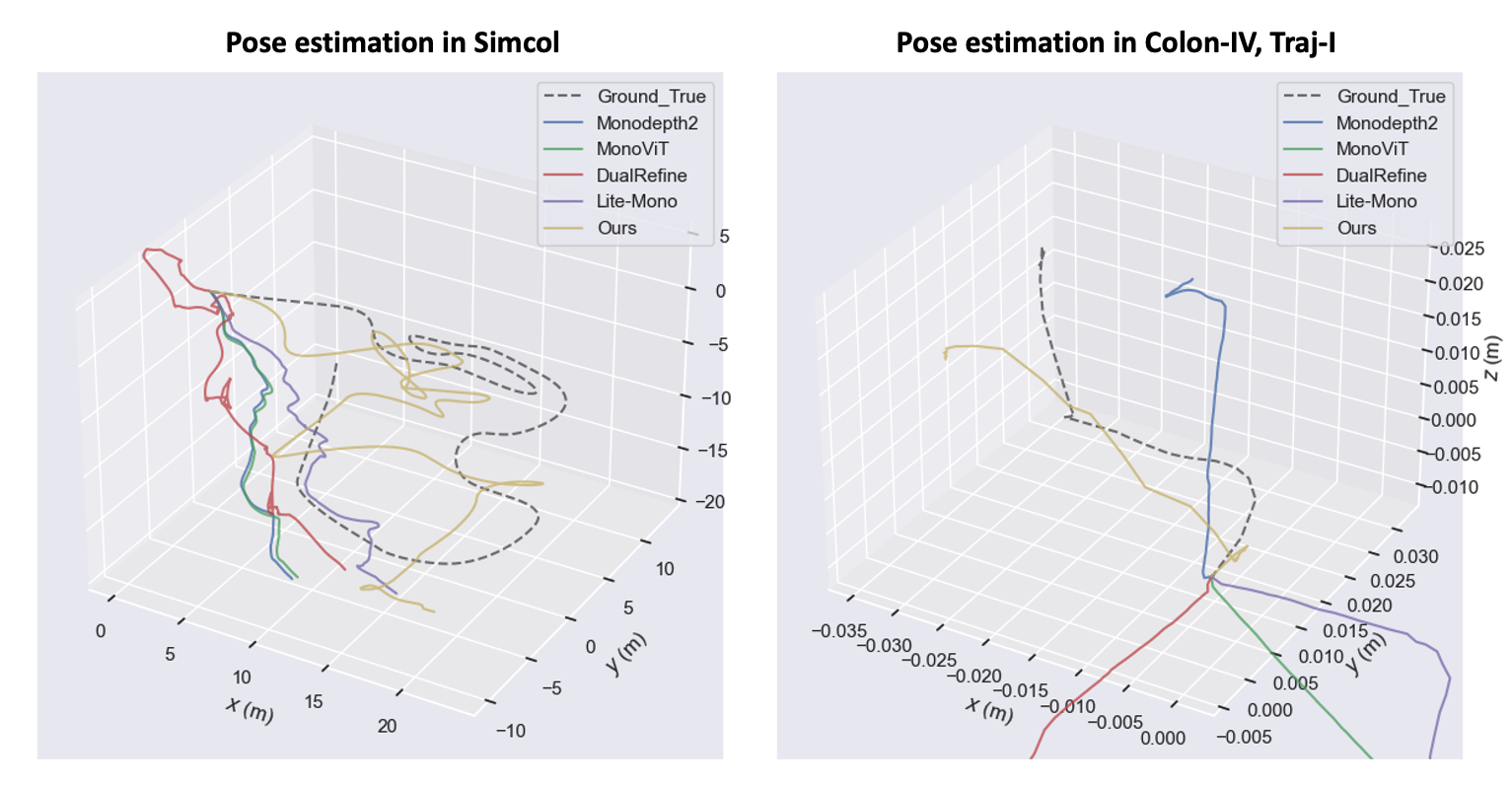}
    \caption{Qualitative results of pose estimation. Our method outperforms others by achieving accurate x, y, and z scale consistency and improved rotation angle alignment.}
    \label{fig:result_pose}
\end{figure}

Existing research often uses synthetic data generated from 3D models, as these provide RGB images with corresponding ground truth depth maps to evaluate endoscopic depth estimation models. Two main datasets are included: the SimCol and EndoSLAM datasets~\cite{rau2024simcol3d,ozyoruk2021endoslam}.

The SimCol dataset, created by Rau et al.~\cite{rau2024simcol3d} using CT scans of the human colon and rendered in Unity, includes 33 scenes with a total of 37,833 RGB frames, each paired with ground truth depth maps and camera poses. Depth values are scaled to [0, 1] representing [0, 20] cm. For evaluation, three trajectories in Synthetic colon I and II are combined as a test set, while Synthetic colon III serves as a test set.

The EndoSLAM dataset~\cite{ozyoruk2021endoslam} includes both synthetic videos generated using the VRCaps simulation environment and real endoscopy videos from porcine GI organs. The synthetic dataset comprises 21,887 colon frames and 12,558 small bowel frames, each with ground truth depth maps and camera poses. These were used as test sets to evaluate model generalization across different GI regions.

To pre-train Generative Latent Bank, we utilized the SceneNet RGB-D dataset~\cite{mccormac2016scenenet}. It contains more than 15,000 frames of synthesized indoor RGB images and corresponding ground truth depth maps.

\subsection{Implementation details}
\textbf{Hyperparameters:} For self-supervised depth and pose estimation, we implemented our method in PyTorch and trained on NVIDIA Quadro RTX 6000 with a batch size of 12, Adam as the optimizer with the weight decay of $1e^{-3}$, an initial learning rate of $1e^{-4}$, input resolution of $480\times480$ and training epoch of 20. To pre-train the generative latent bank, we used a progressive training approach with a fixed learning rate of $10^{-4}$, gradually increasing the output resolution over the course of 100 epochs.

\noindent\textbf{Evaluation metrics:} We report eight commonly used metrics proposed in~\cite{rau2024simcol3d,eigen2014depth} for evaluating the depth estimation accuracy, which are Abs Rel, Sq Rel, RMSE, RMSE log, $\delta < 1.25, \delta < 1.25^2, \delta < 1.25^3$ ~and~$L_1$ error. Relative Translation Error (RTE) and Rotation Error (ROT) are used for evaluating the pose estimation.

\subsection{Results in the SimCol dataset}
\begin{table*}[!t]
\centering
\captionsetup{aboveskip=2pt, belowskip=-10pt} 
\tiny
\renewcommand{\arraystretch}{1.2}
\begin{adjustbox}{width=\textwidth} 
\begin{tabular}{l|l|c|c|c|c|c|c|c|c}
\hline
\textbf{Ablation} & \textbf{Architecture} & \textbf{Abs Rel} & \textbf{Sq Rel} & \textbf{RMSE} & \textbf{RMSE log} & \textbf{$L_1$ error} & \(\delta < 1.25\) & \(\delta < 1.25^2\) & \(\delta < 1.25^3\) \\ 
\hline
\hline
\multirow{4}{*}{\textbf{Simcol-I, II}}
&\textbf{Ours(full)} & \textbf{0.125} & \textbf{0.110} & \textbf{0.614} & \textbf{0.126} & \textbf{0.347} & \textbf{0.937} & \textbf{0.983}  & \textbf{0.996}\\
&Ours w/o Latent Bank & 0.142 & 0.138 & 0.636 & 0.134 & 0.369 & 0.901 & 0.975 & 0.993 \\
&Ours w/o Vae  & 0.129 & 0.112 & 0.627 & 0.128 & 0.365 & 0.920 & 0.983 & 0.995 \\
&Ours w/o Vae and Latent Bank  & 0.151 & 0.167 & 0.660 & 0.142 & 0.381 & 0.899 & 0.968 & 0.987 \\
\hline
\multirow{4}{*}{\textbf{Simcol-III}}
&\textbf{Ours(full)} & \textbf{0.110} & \textbf{0.081} & \textbf{0.505} & \textbf{0.130} & \textbf{0.327} & \textbf{0.924} & \textbf{0.983}  & \textbf{0.991}\\
&Ours w/o Latent Bank & 0.119 & 0.148 & 0.547 & 0.151 & 0.347 & 0.901 & 0.975 & 0.987 \\
&Ours w/o Vae  & 0.114 & 0.112 & 0.522 & 0.135 & 0.329 & 0.906 & 0.977 & 0.990 \\
&Ours w/o Vae and Latent Bank  & 0.126 & 0.130 & 0.562 & 0.149 & 0.375 & 0.896 & 0.964 & 0.988 \\
\hline
\end{tabular}
\end{adjustbox}
\caption{Ablation study on proposed components within our model architecture.}
\label{tab:ablation}
\end{table*}



The proposed framework is evaluated against other representative methods, including Monodepth2~\cite{godard2019digging}, DualRefine~\cite{bangunharcana2023dualrefine}, MonoVit~\cite{zhao2022monovit}, Lite-Mono~\cite{zhang2023lite}, and Depth Pro~\cite{Bochkovskii2024:arxiv}. 
The depth estimation results on the SimCol dataset are presented in Table~\ref{tab:depth}. 
Our method significantly outperforms Monodepth2 across all metrics and demonstrates superior performance compared to other recent methods. 
Specifically, on the Synthetic Colon III test set, our method achieves an RMSE of 0.505, an Abs Rel of 0.11, and an $L_1$ error of 0.327—reducing RMSE by 0.119 and 0.034 compared to Monodepth2 and Lite-Mono, respectively. 
Similarly, our method surpasses all other methods on the Synthetic Colon I/II test set. Depth Pro was also evaluated based on transfer learning (i.e., the pretrained foundation model for natural scene depth estimation is directly used to perform inference on the endoscopy datasets), but the results on both test sets were unsatisfactory.

Fig~\ref{fig:result_depth} presents a qualitative comparison across three different types of endoscopic images (both synthetic datasets and real colonoscopy video frames). Our method consistently delivers high-quality results, even in challenging areas such as reflective surfaces and colonic folds.

Our method demonstrates solid improvement over recent methods in pose estimation, achieving RTE values of 0.089 and 0.296 and ROT values of 0.671 and 0.789 on the Simcol-I/II and Simcol-III test sets, respectively. 
A representative pose trajectory visualization is shown in Fig.~\ref{fig:result_pose} (left), where our method produces a trajectory better aligned with the ground truth compared to other methods. 
In contrast, we observe that many existing methods exaggerate the z-axis scale compared to the x and y axes, resulting in near-linear trajectories that misalign with the colon’s natural curvature.
Colonoscopic movement is challenging to model due to pronounced z-axis progression with gradual curvature at key anatomical points. Our approach addresses this by treating pose outputs as latent variables in a VAE-like framework, regularizing pose transition scales across all axes. This mitigates z-axis dominance, improving sensitivity to x and y axis changes and preserving the natural curvature and spatial alignment of the endoscopic trajectory.

\subsection{Results in the EndoSLAM dataset}
The proposed method is evaluated on the EndoSLAM
dataset to show its generalization ability. 
Table shows the results of compared method on EndoSLAM colon and small intestine datasets. Our proposed method achieves lower RMSE values of 0.54 and 0.503 in colon and small intestine datasets, a relative improvement of 0.369/0.056 and 0.385/0.003 over Monodepth2 and Lite-Mono, respectively. 
We validate compared methods in pose estimation task on synthetic datasets (EndoSLAM Colon and Small Intestine) and porcine colon datasets similar to the human body(EndoSLAMColon-IV, Traj-I). 
In Table~\ref{tab:depth}, we can see that our method maintains superior performance even in non-virtual datasets. 
Fig~\ref{fig:result_pose} (right) shows the pose estimation results in a humanoid colonoscopy clip, where our method is the only one that predicts the correct orientation. Our method’s strength lies in incorporating depth-specific priors via a generative latent bank and VAE-based pose regularization, enhancing robustness in complex endoscopic scenes. This conditioning enables accurate estimation despite challenges like reflective surfaces, colonic folds, and variable lighting, allowing effective generalization across diverse environments without ground truth labels.




\section{Ablation Studies}
\label{sec:ablation}

To further demonstrate the effectiveness of each proposed components in our method, we conducted an ablation study to assess the impact of the generative latent bank and pose VAE modules. 
By selectively removing modules, we evaluated the resulting performance on the SimCol dataset, as shown in Table~\ref{tab:ablation}. 

\textbf{Impact of the Latent Bank Module:} Removing the Latent Bank module led to a marked decrease in all depth estimation metrics, indicating its critical role. 
The generative latent bank effectively conditions the depth output, enhancing robustness against challenging conditions such as reflective surfaces and varying textures. 
This module enables the model to produce consistent and accurate depth maps, even within the self-supervised framework.

\textbf{Impact of the VAE Module:} 
The removal of the VAE module caused a significant drop in pose estimation accuracy, highlighting its critical role in our design. The VAE-constrained PoseNet effectively regulates the scale of x, y, and z-axis transitions, aligning with the inherent characteristics of endoscopic navigation, such as pronounced z-axis progression and occasional erratic movements. This constraint enhances stability yet improves sensitivity in capturing pose changes, preserving alignment with the natural colonoscopic trajectory.
Additionally, removing the VAE module decreases depth estimation metrics, and vice versa. This effect reflects the interdependence of the depth and pose branches in the reprojection process, where both branches contribute to self-supervised learning by warping adjacent frames to predict the current frame, underscoring the interconnected nature of our framework.




\section{Conclusion and Future Directions}
\label{sec:conclusion}

This paper presents a novel architecture for depth and pose estimation in endoscopy, leveraging generative latent priors for robust, self-supervised optimization of both tasks. Our approach achieves state-of-the-art performance on the SimCol and EndoSLAM datasets, showing particular strength in handling complex camera pose variations that challenge existing methods.
Although our results are encouraging, a comprehensive quantitative validation in fully clinical settings remains constrained by the absence of ground truth data from real colonoscopy. In future work, we plan to integrate real-world data through computational methods to approximate ground truth depth and pose, potentially leveraging CT scans or quantitative validation on physical phantoms. Additionally, we aim to incorporate 3D reconstruction of the colon, which will further substantiate our approach and expand its applicability within clinical workflows.

{
    \small
    \bibliographystyle{ieeenat_fullname}
    \bibliography{main}
}


\end{document}